\documentclass{article}
\usepackage{spconf,amsmath,graphicx,hyperref}
\usepackage{amssymb}
\usepackage[table]{xcolor}

\title{BTCCHAT: ADVANCING REMOTE SENSING BI-TEMPORAL CHANGE CAPTIONING WITH MULTIMODAL LARGE LANGUAGE MODEL}
\name{Yujie Li$^{1}$, Wenjia Xu$^{1,\dagger}$, Yuanben Zhang$^{2}$, Zhiwei Wei$^{3}$, Mugen Peng$^{1}$\thanks{$^{\dagger}$Corresponding author: W. Xu, email: xuwenjia@bupt.edu.cn}\thanks{This work has been funded by the National Natural Science Foundation of China under Grant 62301063, and the Young Elite Scientists Sponsorship Program by CAST No.2023QNRC001.}}

\address{$^{1}$State Key Laboratory of Networking and Switching Technology\\Beijing University of Posts and Telecommunications, Beijing, China\\
$^{2}$Aerospace Information Research Institute, Chinese Academy of Sciences, Beijing, China\\
$^{3}$School of Geographical Sciences, Hunan Normal University, Hunan Changsha, China\\}
\begin{document}
\ninept
\maketitle
\begin{abstract}
Bi-temporal satellite imagery supports critical applications such as urbanization monitoring and disaster assessment. Although powerful multimodal large language models~(MLLMs) have been applied in bi-temporal change analysis, previous methods process image pairs through direct concatenation, inadequately modeling temporal correlations and spatial semantic changes. This deficiency hampers visual-semantic alignment in change understanding, thereby constraining the overall effectiveness of current approaches. To address this gap, we propose BTCChat, a multi-temporal MLLM with advanced bi-temporal change understanding capability. BTCChat supports bi-temporal change captioning and retains single-image interpretation capability. To better capture temporal features and spatial semantic changes in image pairs, we design a Change Extraction module. Moreover, to enhance the model's attention to spatial details, we introduce a Prompt Augmentation mechanism, which incorporates contextual clues into the prompt to enhance model performance. Experimental results demonstrate that BTCChat achieves state-of-the-art performance on change captioning and visual question answering tasks. The code is available \href{https://github.com/IntelliSensing/BTCChat}{here}.
\end{abstract}
\section{Introduction}
\label{sec:intro}

\begin{figure*}[tbh!]
  \centering
   \includegraphics[width=0.95\linewidth]{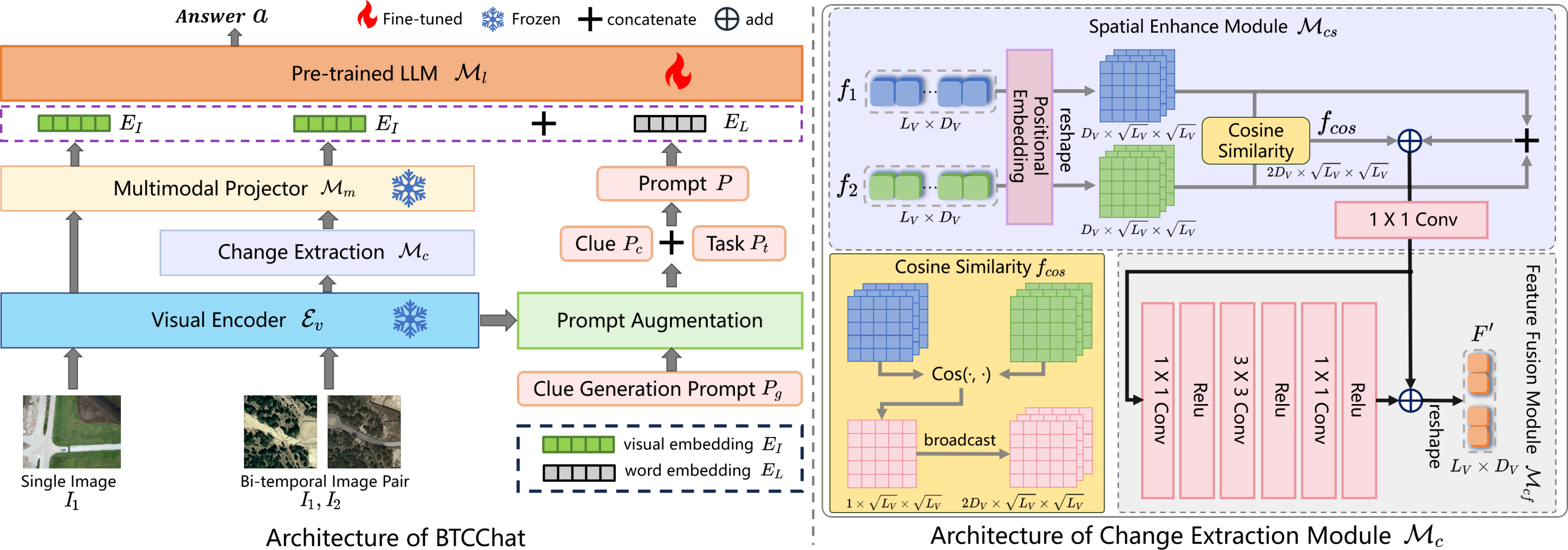}

   \caption{The architecture of our BTCChat. The left part of this figure includes the prompt augmentation mechanism and BTCChat main architecture. The right part of this figure is the structure of the change extraction module.
   }
   \label{fig:framework}
\end{figure*}

With the advancement of satellite remote sensing technologies, a growing volume of bi-temporal image pairs, acquired through satellite revisit imaging, has become available. Unlike single images, these bi-temporal remote sensing image pairs capture temporally correlated information by observing the same geographical location at different time points. By analyzing changes between the two acquisition times, such data can be effectively utilized in a wide range of applications, such as crop monitoring and disaster assessment~\cite{liu2024remote}.

In recent years, deep learning~\cite{qu2025can, xu2023deep, xu2018high} has achieved remarkable success in remote sensing interpretation, where models automatically extract visual features to perform tasks. By coupling visual features with textual world knowledge, multimodal large language models~(MLLMs) exhibit strong visual understanding capabilities and are able to perform a wide range of remote sensing interpretation tasks, demonstrating strong generalization. Building a multi-temporal remote sensing MLLM that supports both single-image and bi-temporal image-pair understanding can greatly enhance the model’s applicability in practical remote sensing interpretation scenarios. Early progress has already been made in applying remote sensing MLLMs to the field of change analysis. ChangeChat~\cite{deng2025changechat} represents an end-to-end MLLM specifically designed for change analysis. 
RingMoGPT~\cite{wang2024ringmogpt} further extends this direction by conducting multi-task training on datasets that include the change captioning task, leading to strong results on this task. Nevertheless, the aforementioned models process bi-temporal image pairs by directly concatenating the visual features of the two images for interpretation, without adequately exploring temporal correlations and fine-grained spatial changes. This limitation hinders visual–semantic alignment in bi-temporal change analysis tasks and leads to deviations in understanding and describing changes.

To fill the gap, we propose BTCChat, a multi-temporal MLLM with advanced \textbf{B}i-\textbf{T}emporal \textbf{C}hange understanding capability. The model is capable of performing change captioning on bi-temporal remote sensing image pairs, while retaining the powerful ability to handle single remote sensing image interpretation. Specifically, to thoroughly capture the temporal correlations and fine-grained spatial changes within bi-temporal image pairs, we design a Change Extraction~(CE) module, which enables the model to uncover the rich temporal semantics embedded in image pairs. In addition, to enhance the model’s attention to spatial details, we propose a Prompt Augmentation~(PA) mechanism, which leverages the prior knowledge of generalist MLLM to preliminarily interpret the visual inputs and enrich task instructions with contextual clues, thereby enhancing the reasoning process of BTCChat. We perform multi-task fine-tuning on a hybrid dataset to train BTCChat and conduct extensive comparative experiments. The results demonstrate that BTCChat achieves state-of-the-art performance on both bi-temporal change captioning and single image visual question answering tasks, highlighting its superior capability in bi-temporal change analysis and generalization on other remote sensing interpretation tasks.

The contributions of our work is as follows:

 \textbf{1)} We propose BTCChat, a multi-temporal MLLM with advanced remote sensing change understanding capability. BTCChat supports change captioning on bi-temporal image pairs and retains powerful single image interpretation capability, thereby advancing the exploration of deep integration between visual features and textual semantics in remote sensing image analysis.
 
 \textbf{2)} We design the Change Extraction~(CE) module to capture fine-grained local spatiotemporal correlation features from bi-temporal image pairs, thereby enhancing the model’s understanding of bi-temporal image pairs. In addition, to enhance the interpretation of spatial details, we introduce the Prompt Augmentation~(PA) mechanism, which leverages generalist MLLM to generate contextual clues that enrich task prompts, further boosting the model’s performance across various tasks.
 
 \textbf{3)} Extensive experiments demonstrate that the BTCChat model achieves state-of-the-art performance on both bi-temporal change captioning and single image visual question answering tasks, showcasing its superior capability in remote sensing image understanding.

\section{METHODOLOGY}
\label{sec:methodology}

In this section, we first introduce the architecture of our proposed BTCChat which includes visual encoder, multimodal projector, LLM, and the proposed Change Extraction~(CE) module, followed by a description of the proposed Prompt Augmentation~(PA) mechanism. Then we introduce our training details.

\subsection{BTCChat Architecture}
\label{sec:BTC_arch}
In the process of remote sensing multimodal tasks, BTCChat receives remote sensing input $I\in \mathbb{R}^{k\times H\times W \times C}$~($k=2$ when input bi-temporal image pairs and $k=1$ when input single images, ) and textual prompt $P$, interprets the visual features, and generates a textual response $a$ following the prompt.

The overall architecture of BTCChat is consistent with mainstream MLLMs, comprising three main components: visual encoder, multimodal projector, and the LLM. To enhance the model’s ability to capture spatiotemporal features in bi-temporal image pairs, we further design the Change Extraction~(CE) module that processes visual features to extract fine-grained spatiotemporal correlations. The overall architecture of the model is illustrated in Fig.~\ref{fig:framework}. In the following sections, we present a detailed description of how BTCChat interprets bi-temporal image pairs.

\textbf{Visual Encoder.} For a given visual input $I \in \mathbb{R}^{2 \times H \times W \times C}$, the visual encoder, formulated as $\mathcal{E}_v$, encodes the visual input $I$ into visual features $F \in \mathbb{R}^{2 \times L_{V} \times D_V}$ ($L_{V}$ refers to the patch number of images and $D_V$ is the dimension of the visual feature):
\begin{equation}
	F = \mathcal{E}_v(I)\,.
\end{equation}
Here, when the visual input is a single image, $I \in \mathbb{R}^{1 \times H \times W \times C}$, the visual feature $F \in \mathbb{R}^{1 \times L_V \times D_V}$ encoded by the visual encoder is directly fed into the multimodal projector.

\textbf{Change Extraction Module.} Achieving advanced performance in change analysis requires models to thoroughly capture temporal features and fine-grained spatial changes within bi-temporal image pairs. Previous approaches that directly concatenate the visual features of the two images are limited for spatiotemporal feature extraction. To enhance change understanding, we design a Change Extraction~(CE) module, formulated as $\mathcal{M}_{c}$, that extracts and reinforces fine-grained spatiotemporal correlations from the visual features $F$.

As is shown in Fig.~\ref{fig:framework}~(right), the CE module consists of the Spatial Enhance module, denoted as $\mathcal{M}_{cs}$, and the Feature Fusion module, denoted as $\mathcal{M}_{cf}$. For the visual features $F=\{f_1, f_2\}\in \mathbb{R}^{2\times L_V \times D_V}$, $\mathcal{M}_c$ extracts and enhances meaningful temporal features and fine-grained spatial changes while preserving the original semantic information.

In the spatial enhancement module $\mathcal{M}_{cs}$, we first introduce learnable positional embeddings to preserve spatial location awareness. Then, $f_1$ and $f_2$ are reshaped into $D_V \times \sqrt[]{L_V} \times \sqrt[]{L_V}$. Subsequently, their cosine similarity embedding, $f_{cos}$, is calculated and added to the concatenated visual features $[f_1;f_2]$, thereby enhancing the local spatial change correlations between the bi-temporal visual features.
\begin{equation}
\begin{aligned}
   & f_{cos}=\Uparrow (Cos(f_1, f_2)) \,, \\
   & [f_1';f_2'] = [f_1;f_2] + f_{cos}\,.
\end{aligned}
\end{equation}
Here, $[\cdot;\cdot]$ denotes concatenation, $Cos$ denotes calculating cosine similarity, and $\Uparrow$ denotes broadcast. $[f_1;f_2] \in \mathbb{R}^{2D_V \times \sqrt{L_V} \times \sqrt{L_V}}$ and we broadcast the cosine similarity from $1 \times \sqrt{L_V} \times \sqrt{L_V}$ to $2D_V \times \sqrt{L_V} \times \sqrt{L_V}$.

Subsequently, in the feature fusion module $\mathcal{M}_{cf}$, we design a three-layer 2D convolutional network with kernel sizes of $1\times1$, $3\times3$, and $1\times1$, respectively. ReLU is the activation function between layers, and residual connections are introduced. We fuse the visual features, extracting local spatiotemporal features where changes occur, and then enhance them upon the global visual features through the multi-layer 2D convolutional network:
\begin{equation}
    F' = Conv_1([f_1',f_2']) +Conv_m([f_1',f_2'])\,,
\end{equation}
where $F' \in \mathbb{R}^{D_V \times \sqrt{L_V} \times \sqrt{L_V}}$ and is then reshaped into $L_V \times D_V$, $Conv_m$ denotes the multi-layer convolution network, $Conv_1$ denotes a single layer convolution with the kernel size of $1 \times 1$, halving the feature dimension to $D_V$.

\textbf{Multimodal Projector.} The CE module encodes the visual feature $F$ into enhanced feature $F'$. The multimodal projector, formulated as $\mathcal{M}_m$, maps the feature from the visual feature space to the word embedding space with $1/4$ downsampling which helps extract high-level semantics while significantly reducing computation cost:
\begin{equation}
    E_I = \mathcal{M}_m(F')\,.
\end{equation}
Here, $E_I \in \mathbb{R}^{L_d \times D_L}$~($L_d=L_V/4$ and $D_L$ is the dimension of word embedding).

\textbf{Large Language Model.} The visual embedding $E_I$ is concatenated with the word embedding, formulated as $E_L$, which is obtained from tokenizing the prompt $P$, to get the multimodal embedding. The LLM then generates a response $a$ corresponding to the multimodal embedding:
\begin{equation}
    a = \mathcal{M}_{l}([E_I;E_L])  \,.
\end{equation}

\subsection{Prompt Augmentation Mechanism}
\label{sec:PA_mech}
For effective bi-temporal change analysis, the model not only needs to thoroughly capture the spatiotemporal correlations within image pairs but also attend to the spatial details of each individual image. To this end, we propose the Prompt Augmentation~(PA) mechanism. 

The PA mechanism utilizes the frozen base model of BTCChat~(to maintain consistency in text generation style) to provide the descriptions of the input remote sensing images, which serves as contextual clues for BTCChat when performing remote sensing tasks. This mechanism is designed based on the fundamental understanding capabilities of generalist MLLMs for remote sensing images and we utilize it on both bi-temporal and single-image understanding tasks.

In the Prompt Augmentation mechanism, when the model performs remote sensing interpretation tasks, the user provides a visual input $I$ and a textual task instruction $P_t$. The visual input $I$ is first processed by the frozen base model, denoted as $\mathcal{M}_b$. Guided by a pre-defined prompt $P_g$, the base model generates descriptions for each input image. These descriptive texts, denoted as $P_c$, serve as contextual clues, which are concatenated with the task instruction $P_t$ using a fixed template to form the final prompt $P$:
\begin{equation}
\begin{aligned}
   & P_{c} = \mathcal{M}_{b}(I,P_{g})\,, \\
   & P = Template(P_{c}, P_{t})\,.
\end{aligned}
\end{equation}
Our proposed BTCChat then receives the visual input $I$ and the textual prompt $P$, and performs the remote sensing interpretation tasks. Here, we set the $P_g$ as ``Please describe the remote sensing image(s) in detail''.

The responses $P_{c}$ generated by $\mathcal{M}_{b}$ contain image details that serve as clues, enhancing the model's perception of spatial details, which in turn improves task performance. By adding these clues to the $P$, the Prompt Augmentation mechanism not only provides contextual clues to support the inference of BTCChat without introducing additional training but also enriches the instructions, thus augmenting the prompt for the model.

\subsection{Training Details}
\label{sec:train_detail}
Our goal is to train a remote sensing MLLM with strong bi-temporal change understanding capabilities, achieving superior change captioning performance while preserving its ability to interpret single-image remote sensing inputs. To this end, we employ the LEVIR-CC~\cite{liu2022remote} and GeoChat-Instruct~\cite{kuckreja2024geochat} datasets for multi-task instruction tuning. LEVIR-CC is a large-scale, manually annotated change captioning dataset comprising 10,077 bi-temporal image pairs, each associated with five captions. We use its training split and construct instruction–answer pairs based on the original annotations for model training. GeoChat-Instruct, on the other hand, is a large-scale single-image remote sensing instruction tuning dataset covering a variety of tasks, including scene classification, visual question answering, and referring expression, thereby equipping the model with strong single-image interpretation capabilities.

BTCChat adopts VILA-1.5~(3B)~\cite{lin2024vila}, a robust and powerful MLLM, as its base model, with the addition of the proposed CE module. We conduct full-parameter fine-tuning using the AdamW optimizer with a cosine learning rate scheduler, a warmup ratio of 3\%, and a batch size of 128. The loss function is the cross-entropy loss. To balance the training of different modules, we design a two-stage training strategy. In the first stage, we train only the CE module on the LEVIR-CC dataset for the change captioning task, unfreezing only the CE module, with a learning rate of $1e^{-3}$ for 10 epochs. In the second stage, we unfreeze only the LLM module and perform multi-task joint training on GeoChat-Instruct and LEVIR-CC, with a learning rate of $1e^{-4}$ for 1 epoch. Both training stages are executed on four NVIDIA RTX 4090 GPUs.

\section{EXPERIMENTS}
\label{sec:experiments}

In this section, we evaluate BTCChat on bi-temporal change captioning and single-image visual question answering tasks, and further conduct ablation studies on the Change Extraction module and Prompt Augmentation mechanism.

\subsection{Change Captioning}
\label{sec:changecaption}

\begin{table}[tb!]
\centering
\caption{Comparison with specialist models and MLLMs on the change captioning task.}
\label{tab: cc}
\renewcommand{\arraystretch}{1.2}\fontsize{9pt}{11pt}\selectfont\resizebox{1\linewidth}{!}{
\begin{tabular}{lcccc}
\hline
\textbf{Method}            & \textbf{BLEU-1} & \textbf{METEOR} & \textbf{ROUGE-L} & \textbf{CIDEr-D} \\ \hline
\textbf{Specialist models} &                 &                 &                  &                  \\
RSICCFormer~\cite{liu2022remote}                & 81.96           & 38.16           & 72.57            & 132.00           \\
PSNet~\cite{liu2023progressive}                      & 83.86           & 38.80           & 73.60            & 132.62           \\
PromptCC~\cite{liu2023decoupling}                   & 83.66           & 38.82           & 73.72            & 136.44           \\
Chg2Cap~\cite{chang2023changes}                    & 86.14           & 40.03           & 75.12            & 136.61           \\
MADiffCC~\cite{yang2024remote}                   & \textbf{86.28}  & 40.16           & \textbf{75.37}   & \underline{138.61}     \\ \hline
\textbf{RS MLLMs}          &                 &                 &                  &                  \\
RingMoGPT~\cite{wang2024ringmogpt}                  & 83.26           & \underline{40.25}  & 73.97            & 135.32           \\
ChangeChat~\cite{deng2025changechat}                 & 83.14           & 38.73           & 74.01            & 136.56           \\
TEOChat~\cite{irvin2024teochat}   &  84.56  &  40.11  &  74.44  &     137.49  \\
\rowcolor[HTML]{fce0dd}
OURS~(w/o CE)  &  81.62  &  37.99  &  72.41   &   131.28  \\
\rowcolor[HTML]{fce0dd}
OURS~(w/o PA)  &  84.49  &  39.45  &  74.96   &   138.47  \\
\rowcolor[HTML]{fce0dd}
OURS                       & \underline{86.20}     & \textbf{40.31}     & \underline{75.15}      & \textbf{139.12}  \\
\hline
\end{tabular}}
\end{table}

We evaluate the change captioning capability of BTCChat on the test set of LEVIR-CC~\cite{liu2022remote}, comparing it against both specialist models~(i.e., RSICCFormer~\cite{liu2022remote}, PSNet~\cite{liu2023progressive}, PromptCC~\cite{liu2023decoupling}, Chg2Cap~\cite{chang2023changes}, and MADiffCC~\cite{yang2024remote}) and other remote sensing MLLMs~(i.e., RingMoGPT~\cite{wang2024ringmogpt}, ChangeChat~\cite{deng2025changechat}, and TEOChat~\cite{irvin2024teochat}). Specifically, TEOChat, an MLLM with temporal understanding capabilities, is fine-tuned on the LEVIR-CC training set for 5 epochs based on its publicly released checkpoint before evaluation. 

\begin{figure}[tb!]
  \centering
   \includegraphics[width=1\linewidth]{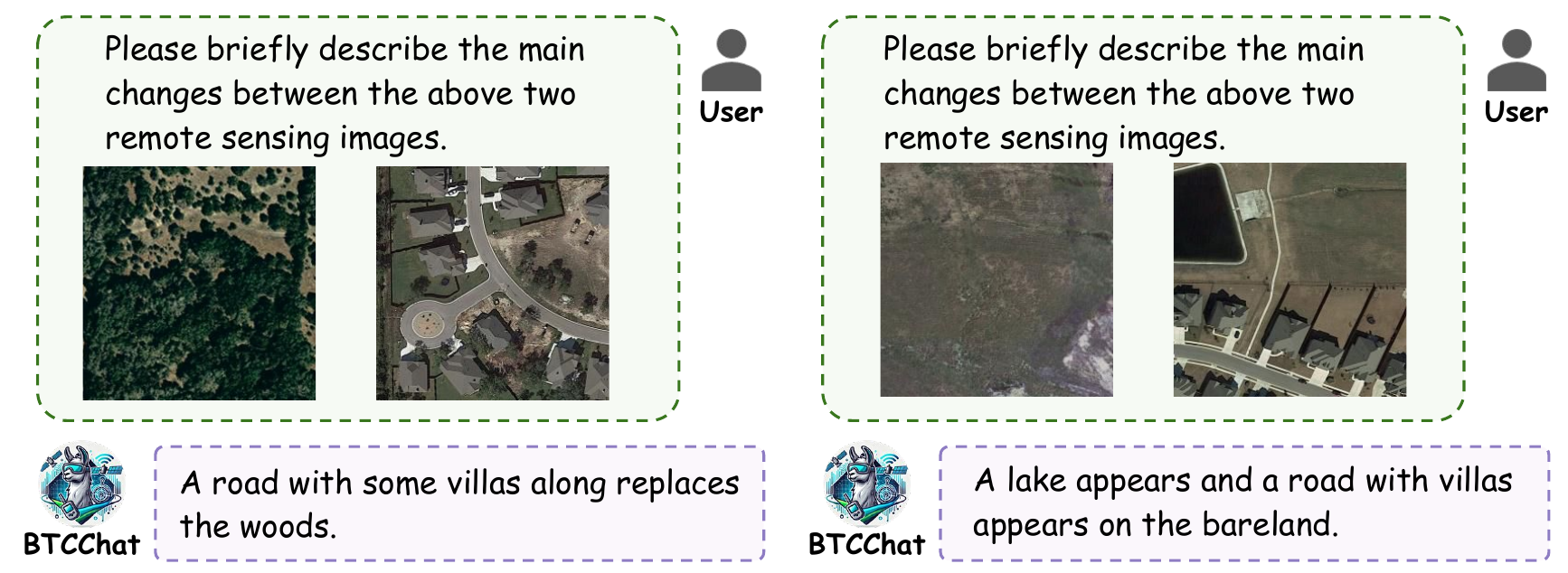}
   \caption{Qualitative visualization of BTCChat's performance on the change captioning task.}
   \label{fig:cc}
\end{figure}

Experimental results show that BTCChat achieves either state-of-the-art or comparable performance across all metrics for the change captioning task. As shown in Table~\ref{tab: cc}, BTCChat surpasses specialist models on the METEOR and CIDEr-D metrics, reaching 40.31 and 139.12 respectively and achieving sota performance, while achieving results on BLEU-1 and ROUGE-L that are on par with the best-performing models. Moreover, BTCChat consistently outperforms other MLLMs across all metrics. These improvements can be attributed to the CE module, which is specifically designed to extract spatiotemporal correlation features from bi-temporal image pairs. We present two illustrative examples of BTCChat performing the change captioning task in Fig.~\ref{fig:cc}, where the model effectively identifies and describes the observed changes.

\subsection{Visual Question Answering}
\label{sec:visualqa}

\begin{table}[tb!]
\centering
\caption{Comparison with specialist models and RS MLLMs on the visual question answering task using the test set of RSVQA-LR.}
\label{tab: lr}
\renewcommand{\arraystretch}{1.2}\fontsize{9pt}{11pt}\selectfont\resizebox{\linewidth}{!}{
\begin{tabular}{lcccc}
\hline
\textbf{Method} & \textbf{Presence} & \textbf{Comparison} & \textbf{Rural/Urban} & \textbf{Avg. Accuracy} \\ \hline
\textit{\textbf{Specialist models}} &             &             &                &             \\
RSVQA~\cite{lobry2020rsvqa}                               & 87.47       & 81.50       & 90.00          & 86.32       \\
Bi-Modal~\cite{bazi2022bi}                            & 91.06       & 91.16 & \underline{92.66}    & 91.63       \\
SHRNet~\cite{zhang2023spatial}                              & 91.03       & 90.48       & 94.00          & 91.84 \\
RSGPT~\cite{hu2025rsgpt}           & 91.17    & \underline{91.70}      & 94.00                & \textbf{92.29}         \\ \hline
\textit{\textbf{RS MLLMs}}     &             &             &                &             \\
SkyEyeGPT~\cite{zhan2025skyeyegpt}                           & 88.93       & 88.63       & 75.00          & 84.19       \\
LHRS-Bot~\cite{muhtar2024lhrs}                            & 89.07       & 88.51       & 90.00          & 89.19       \\
VHM~\cite{pang2025vhm}                                 & 90.11       & 89.89       & 88.00          & 89.33       \\
GeoChat~\cite{kuckreja2024geochat}                             & 91.09 & 90.33       & \textbf{94.00} & 90.70       \\
\rowcolor[HTML]{fce0dd}
OURS~(w/o PA)  &  \underline{91.51}  &  91.85  &  92.00  &  91.71  \\
\rowcolor[HTML]{fce0dd}
OURS                                & \textbf{91.64}       & \textbf{92.68}       & 90.00          & \underline{92.21}       \\ 
\hline
\end{tabular}}
\end{table}

\begin{table}[tb!]
\centering
\caption{Zero-shot evaluation on the visual question answering task using the test set 2 of RSVQA-HR.}
\label{tab: hr}
\renewcommand{\arraystretch}{1.2}\fontsize{9pt}{11pt}\selectfont\resizebox{0.95\linewidth}{!}{
\begin{tabular}{lccc}
\hline
\textbf{Method}                 & \textbf{Presence} & \textbf{Comparison} & \textbf{Avg. Accuracy} \\ \hline
\textit{\textbf{Generalist MLLMs}} &                   &                     &                        \\
MiniGPTv2~\cite{chen2023minigpt}                       & 40.79             & 50.91               & 46.46                  \\
LLaVA-1.5~\cite{liu2024visual}                       & \textbf{68.23}    & 65.45               & 66.67                  \\ \hline
\textit{\textbf{RS MLLMs}}      &                   &                     &                        \\
GeoChat~\cite{kuckreja2024geochat}                         & 59.02             & 83.16      & \underline{72.53}         \\
EarthGPT~\cite{zhang2024earthgpt}                        & \underline{62.77}       & 79.53               & 72.06                  \\
\rowcolor[HTML]{fce0dd}
OURS~(w/o PA)  &  56.95  &  \underline{83.91}  &  72.21  \\
\rowcolor[HTML]{fce0dd}
OURS                            & 59.29             & \textbf{84.05}         & \textbf{73.15}            \\ 
\hline
\end{tabular}}
\end{table}

We further evaluate the capability of BTCChat on the single-image visual question answering task using RSVQA-LR and RSVQA-HR~\cite{lobry2020rsvqa} datasets.

On RSVQA-LR, BTCChat is compared against both specialist models~(i.e., RSVQA~\cite{lobry2020rsvqa}, Bi-Modal~\cite{bazi2022bi}, SHRNet~\cite{zhang2023spatial}, and RSGPT~\cite{hu2025rsgpt}) and remote sensing MLLMs~(i.e., SkyEyeGPT~\cite{zhan2025skyeyegpt}, LHRS-Bot~\cite{muhtar2024lhrs}, VHM~\cite{pang2025vhm}, and GeoChat~\cite{kuckreja2024geochat}). Here, the RS MLLMs are all trained on supersets of RSVQA-LR. As shown in Table~\ref{tab: lr}, BTCChat achieves performance second only to the best specialist model, with an average accuracy of 92.21\%, which is comparable to the 92.29\% obtained by RSGPT. Moreover, BTCChat attains state-of-the-art accuracy on the ``Presence'' and ``Comparison'' question types. 

On RSVQA-HR, we conduct zero-shot testing, comparing BTCChat with both generalist MLLMs~(i.e., MiniGPTv2~\cite{chen2023minigpt} and LLaVA-1.5~\cite{liu2024visual}) and remote sensing MLLMs~(GeoChat~\cite{kuckreja2024geochat} and EarthGPT~\cite{zhang2024earthgpt}). The results, presented in Table~\ref{tab: hr}, show that BTCChat achieves the best performance, with an average accuracy of 73.15\%, surpassing GeoChat. These results demonstrate that BTCChat retains strong capabilities for understanding single remote sensing images, a benefit attributable to the incorporation of GeoChat-Instruct into the joint training process.

\subsection{Ablation Studies}
\label{sec:ablations_ce_pa}
We conduct ablation studies on the Change Extraction~(CE) module and Prompt Augmentation~(PA) mechanism, which validate the effectiveness of their designs.

\textbf{Ablation Study on CE Module.} We perform an ablation study on the CE module on the change captioning task. In the control setting, the CE module is removed, and the visual features of the two images in each pair are directly concatenated as input, while all other training strategies and datasets remain unchanged. As shown in Table~\ref{tab: cc}~(w/o CE), incorporating the CE module yields a significant performance improvement on the CIDEr-D metric, increasing from 131.28 to 139.12, along with consistent gains across other metrics. These results demonstrate that the CE module effectively enhances the model’s ability to understand bi-temporal changes.

\textbf{Ablation Study on PA Mechanism.} We conduct ablation studies on the PA mechanism on both change captioning and visual question answering tasks. In the control setting, user instructions are directly fed into the model with other settings unchanged. The results, presented in Table~\ref{tab: cc},~\ref{tab: lr}, and~\ref{tab: hr}~(w/o PA), demonstrate that the PA mechanism effectively improves BTCChat’s performance on both tasks. Notably, in the visual question answering task, the improvement is most pronounced on the ``Presence'' question type in RSVQA-HR, where accuracy increases from 56.95\% to 59.29\%. The average accuracy also rises from 72.21\% to 73.15\%. These results indicate that the PA mechanism effectively enhances task performance by introducing contextual clues into the prompt.

\section{CONCLUSION}
In this work, we propose BTCChat, a multi-temporal MLLM with advanced capability for remote sensing change understanding. The model is capable of performing both bi-temporal change captioning and single-image interpretation tasks. To strengthen its ability to capture fine-grained spatiotemporal correlations, we design a Change Extraction (CE) module that explicitly models inter-temporal differences. In addition, to enhance the model's attention on the spatial details, we introduce the Prompt Augmentation mechanism that incorporates contextual clues about image details into prompts to further improve model performance. Extensive comparative experiments demonstrate that BTCChat achieves state-of-the-art or comparable performance on both change captioning and visual question answering tasks.




\vfill\pagebreak




\bibliographystyle{IEEEbib}
\bibliography{strings,refs}

\end{document}